\documentclass[conference]{IEEEtran}
\IEEEoverridecommandlockouts
\usepackage{cite}
\usepackage{amsmath,amssymb,amsfonts}
\usepackage{algcompatible}
\usepackage{graphicx}
\usepackage{textcomp}
\usepackage{xcolor}
\usepackage{algorithm}
\usepackage{algpseudocode}
\usepackage{amsmath}
\usepackage{amsfonts}
\usepackage{amssymb} 
\usepackage{multirow}
\usepackage{booktabs}
\usepackage{graphicx}
\usepackage{array}  
\usepackage{amsmath}  
\def\BibTeX{{\rm B\kern-.05em{\sc i\kern-.025em b}\kern-.08em
    T\kern-.1667em\lower.7ex\hbox{E}\kern-.125emX}}
\begin{document}

\title{EM-MIAs: Enhancing Membership Inference Attacks in Large Language Models through Ensemble Modeling
}

\author{\IEEEauthorblockN{Zichen Song}
\IEEEauthorblockA{\textit{School of Information Science and} \\
\textit{Engineering} \\
\textit{Lanzhou University}\\
Lanzhou, China \\
songzch21@lzu.edu.cn}
\and
\IEEEauthorblockN{Sitan Huang}
\IEEEauthorblockA{\textit{School of Information Science and} \\
\textit{Engineering} \\
\textit{Lanzhou University}\\
Lanzhou, China \\
320220937820@lzu.edu.cn}
\and
\IEEEauthorblockN{Zhongfeng Kang*}
\IEEEauthorblockA{\textit{School of Information Science and} \\
\textit{Engineering} \\
\textit{Lanzhou University}\\
Lanzhou, China \\
kangzf@lzu.edu.cn}}

\maketitle

\begin{abstract}
With the widespread application of large language models (LLM), concerns about the privacy leakage of model training data have increasingly become a focus. Membership Inference Attacks (MIAs) have emerged as a critical tool for evaluating the privacy risks associated with these models. Although existing attack methods, such as LOSS, Reference-based, min-k, and zlib, perform well in certain scenarios, their effectiveness on large pre-trained language models often approaches random guessing, particularly in the context of large-scale datasets and single-epoch training. To address this issue, this paper proposes a novel ensemble attack method that integrates several existing MIAs techniques (LOSS, Reference-based, min-k, zlib) into an XGBoost-based model to enhance overall attack performance (EM-MIAs). Experimental results demonstrate that the ensemble model significantly improves both AUC-ROC and accuracy compared to individual attack methods across various large language models and datasets. This indicates that by combining the strengths of different methods, we can more effectively identify members of the model's training data, thereby providing a more robust tool for evaluating the privacy risks of LLM. This study offers new directions for further research in the field of LLM privacy protection and underscores the necessity of developing more powerful privacy auditing methods.
\end{abstract}


\section{Introduction} The rapid advancement of large language models (LLMs) has transformed a wide array of applications across diverse fields, including natural language processing, machine translation, content generation, and more. These models, often trained on massive datasets, have shown impressive capabilities in understanding and generating human-like text. However, the extensive deployment of LLMs has also brought to light critical concerns regarding the potential leakage of private or sensitive information embedded in the training data. One of the key techniques used to assess the privacy risks associated with machine learning models, particularly LLMs, is Membership Inference Attacks (MIAs). These attacks attempt to determine whether a specific data point was used in training the model, thereby exposing potential privacy vulnerabilities that could lead to unintended information leaks. [1-6]

Over the years, several Membership Inference Attack techniques have been introduced, with notable methods including LOSS-based attacks, Reference-based approaches, min-k, and zlib entropy techniques. Each of these methods exploits different aspects of the model's behavior to infer membership status. Although these techniques have demonstrated effectiveness in specific contexts, their performance often declines when applied to large pre-trained language models. This decline is especially evident in large-scale datasets and scenarios where models undergo only single-epoch training, conditions under which LLMs tend to generalize more effectively. As a result, traditional MIAs frequently yield results close to random guessing, reducing their utility in practical applications involving large-scale LLMs. This limitation highlights the challenge of applying conventional MIAs to modern LLMs, as these models are specifically designed to minimize overfitting and exhibit strong generalization capabilities, which in turn reduces the exploitable signals that MIAs typically rely on for accurate membership inference. [7-14]

To address the inherent limitations of existing approaches, this paper proposes a novel ensemble strategy that combines multiple MIA methods into a single unified framework using XGBoost. This ensemble model, referred to as EM-MIAs, aims to leverage the complementary strengths of individual attack methods to enhance the overall accuracy and robustness of membership inference against large language models. By aggregating the outputs of different MIAs techniques, we hypothesize that the ensemble model can detect subtle patterns that may go unnoticed by individual methods, thereby providing more precise and reliable membership inference. Through extensive experimentation on a variety of large language models and diverse datasets, we demonstrate that the proposed ensemble model significantly outperforms traditional MIAs techniques in terms of AUC-ROC and accuracy, offering a more effective tool for evaluating privacy risks in LLMs.[15-23]

Our main contributions are as follows:

\begin{figure*}[t]
    \centering
    \includegraphics[width=\textwidth]{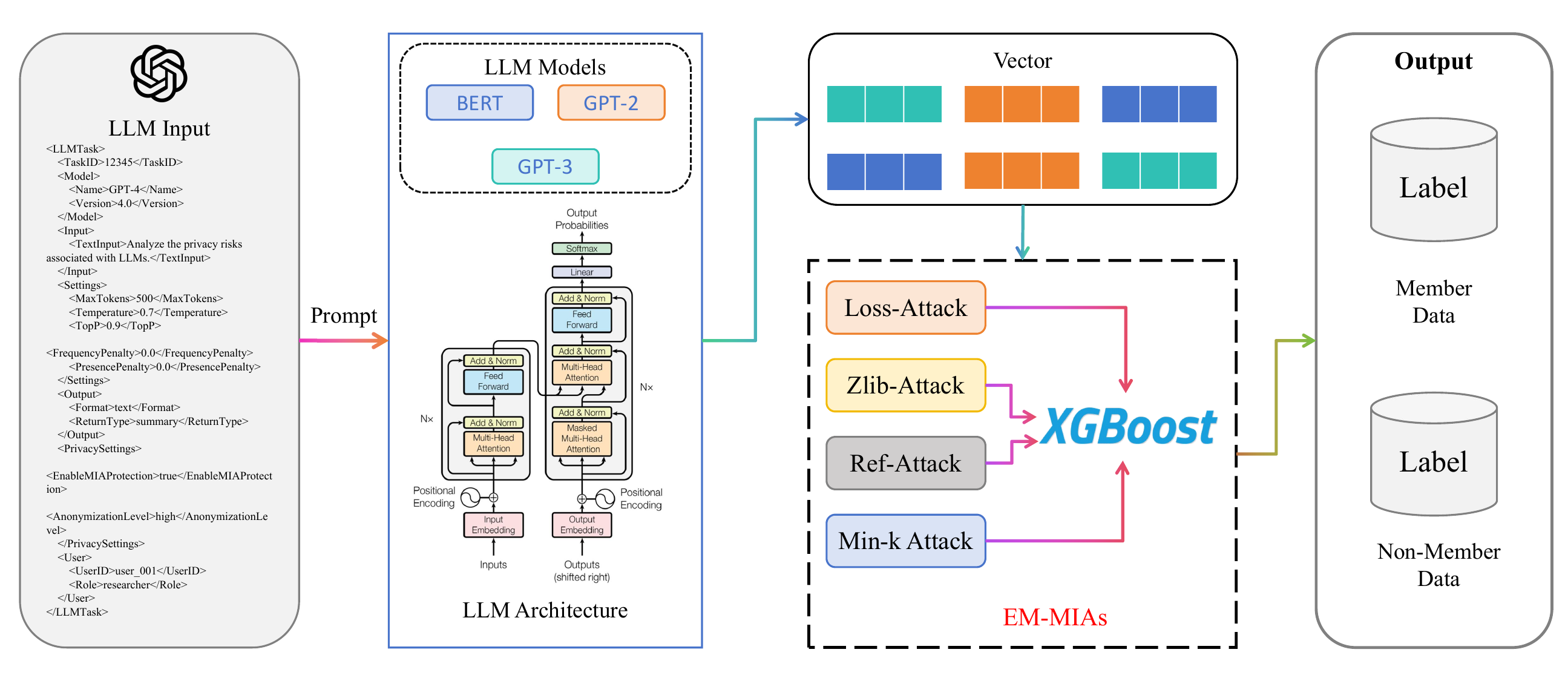}
    \caption{The diagram illustrates the LLM architecture and the process flow of various attack methods, including Loss-Attack, Zlib-Attack, Ref-Attack, and Min-k Attack, which use an XGBoost classifier to distinguish between member and non-member data.(EM-MIAs)}
    \label{fig:LLM_MIA}
\end{figure*}

\begin{itemize}
    \item \textbf{Proposed a Novel Ensemble Model:} We introduce an ensemble model that integrates multiple existing MIAs methods using XGBoost to enhance membership inference on large language models.
    \item \textbf{Comprehensive Evaluation:} We conduct extensive experiments on various large language models and datasets, demonstrating that our ensemble model significantly outperforms individual MIAs methods in terms of AUC-ROC and accuracy. 
    \item \textbf{Insights into Privacy Risks:} Our findings provide new insights into the privacy risks associated with LLM, highlighting the need for more sophisticated privacy-preserving techniques.
\end{itemize}

\section{Methodology}
We propose a novel ensemble approach for Membership Inference Attacks (MIAs) on Large Language Models (LLMs), integrating multiple MIA techniques—LOSS, Reference-based, Min-k\%, and zlib entropy—using an ensemble learning framework called EM-MIAs. This method leverages the strengths of each individual attack to improve the robustness and accuracy of MIAs in LLMs by combining various strategies that capture different model behaviors and vulnerabilities. [24-35]

\subsection{Problem Definition}
Given a pre-trained LLM $\mathcal{M}$, a training dataset $\mathcal{D}_{train}$, and a target data point $x \in \mathcal{D}_{test}$ (where $\mathcal{D}_{test}$ is independent of $\mathcal{D}_{train}$), the goal of MIAs is to infer if $x$ was in $\mathcal{D}_{train}$. The attack model $\mathcal{A}$ outputs a binary decision $\hat{y} \in \{0,1\}$:

\[
\hat{y} = 
\begin{cases} 
1 & \text{if } x \in \mathcal{D}_{train}, \\
0 & \text{otherwise}.
\end{cases}
\]

The performance of the attack model $\mathcal{A}$ is typically evaluated using the AUC-ROC metric.

\subsection{Overview of Individual MIA Methods}

\textbf{LOSS Attack:} This method computes the model's prediction loss for $x$. Training data points often exhibit lower loss than non-training points. Loss $L(x)$ is calculated using cross-entropy:

\[
L(x) = -\sum_{i=1}^{C} y_i \log \hat{y}_i
\]

where $C$ is the number of classes, $y_i$ is the true label, and $\hat{y}_i$ is the predicted probability.

\textbf{Reference-based Attack:} This attack compares the loss of $x$ with that from a reference model $\mathcal{M}_{ref}$ trained on a disjoint dataset. The feature is the loss difference $\Delta L(x) = L(x) - L_{ref}(x)$, highlighting discrepancies between the target and reference models.

\textbf{Min-k\% Attack:} This approach focuses on the k\% of tokens in $x$ with the lowest likelihood under the model’s predictions, computing the average likelihood for these tokens:

\[
\text{Min-k\%}(x) = \frac{1}{k} \sum_{j \in \text{Min-k\%}} \hat{p}_j
\]

\textbf{zlib Entropy Attack:} Normalizes the loss of $x$ by its zlib compression size $Z(x)$, adjusting for data complexity:

\begin{table*}[t]
\caption{AUC ROC of Membership Inference Attacks (MIAs) on Different Datasets. The table highlights the effectiveness of the EM method across various datasets.}
\centering
\resizebox{0.8\textwidth}{!}{%
\begin{tabular}{lcccccc}
\toprule
\textbf{DataSets} & \textbf{Params} & \textbf{LOSS} & \textbf{Ref} & \textbf{min-k} & \textbf{zlib} & \textbf{EM-MIAs} \\
\midrule
\multirow{5}{*}{\centering Wikipedia} & 160M & 0.505±0.003 & 0.517±0.004 & 0.490±0.002 & 0.517±0.004 & 0.626±0.003 \\
                                      & 1.4B & 0.511±0.002 & 0.545±0.003 & 0.508±0.003 & 0.520±0.002 & 0.629±0.004 \\
                                      & 2.8B & 0.517±0.003 & 0.567±0.004 & 0.514±0.002 & 0.526±0.003 & 0.631±0.002 \\
                                      & 6.9B & 0.515±0.004 & 0.572±0.003 & 0.515±0.002 & 0.519±0.004 & 0.636±0.003 \\
                                      & 12B  & 0.518±0.003 & 0.580±0.004 & 0.520±0.003 & 0.525±0.002 & 0.632±0.003 \\
\midrule
\multirow{5}{*}{\centering Github}    & 160M & 0.506±0.004 & 0.519±0.003 & 0.536±0.002 & 0.523±0.003 & 0.693±0.004 \\
                                      & 1.4B & 0.513±0.003 & 0.545±0.002 & 0.556±0.004 & 0.531±0.003 & 0.773±0.002 \\
                                      & 2.8B & 0.517±0.002 & 0.552±0.004 & 0.568±0.003 & 0.543±0.003 & 0.810±0.003 \\
                                      & 6.9B & 0.517±0.003 & 0.564±0.003 & 0.575±0.002 & 0.557±0.004 & 0.786±0.004 \\
                                      & 12B  & 0.517±0.002 & 0.575±0.004 & 0.580±0.003 & 0.570±0.002 & 0.792±0.004 \\
\midrule
\multirow{5}{*}{\centering Pile CC}   & 160M & 0.500±0.003 & 0.504±0.002 & 0.498±0.004 & 0.501±0.003 & 0.597±0.002 \\
                                      & 1.4B & 0.501±0.002 & 0.526±0.003 & 0.510±0.002 & 0.502±0.004 & 0.602±0.003 \\
                                      & 2.8B & 0.504±0.004 & 0.538±0.003 & 0.510±0.003 & 0.505±0.002 & 0.598±0.003 \\
                                      & 6.9B & 0.507±0.002 & 0.548±0.004 & 0.514±0.002 & 0.514±0.003 & 0.605±0.004 \\
                                      & 12B  & 0.512±0.003 & 0.583±0.002 & 0.519±0.003 & 0.515±0.004 & 0.623±0.003 \\
\midrule
\multirow{5}{*}{\centering PubMed Central} & 160M & 0.503±0.002 & 0.519±0.004 & 0.506±0.003 & 0.507±0.002 & 0.606±0.004 \\
                                           & 1.4B & 0.507±0.003 & 0.533±0.002 & 0.532±0.003 & 0.532±0.004 & 0.633±0.002 \\
                                           & 2.8B & 0.499±0.004 & 0.538±0.002 & 0.537±0.004 & 0.537±0.003 & 0.638±0.002 \\
                                           & 6.9B & 0.507±0.003 & 0.553±0.004 & 0.553±0.003 & 0.554±0.002 & 0.654±0.003 \\
                                           & 12B  & 0.504±0.004 & 0.562±0.002 & 0.561±0.003 & 0.560±0.004 & 0.661±0.002 \\
\midrule
\multirow{5}{*}{\centering ArXiv}     & 160M & 0.509±0.003 & 0.510±0.002 & 0.499±0.004 & 0.508±0.002 & 0.608±0.003 \\
                                      & 1.4B & 0.515±0.004 & 0.516±0.002 & 0.513±0.003 & 0.519±0.002 & 0.623±0.004 \\
                                      & 2.8B & 0.520±0.003 & 0.523±0.004 & 0.520±0.002 & 0.526±0.003 & 0.640±0.002 \\
                                      & 6.9B & 0.523±0.002 & 0.539±0.003 & 0.523±0.004 & 0.539±0.002 & 0.657±0.003 \\
                                      & 12B  & 0.529±0.003 & 0.557±0.002 & 0.527±0.004 & 0.557±0.003 & 0.660±0.002 \\
\midrule
\multirow{5}{*}{\centering DM Math}   & 160M & 0.500±0.002 & 0.507±0.003 & 0.495±0.004 & 0.484±0.002 & 0.585±0.003 \\
                                      & 1.4B & 0.507±0.004 & 0.513±0.002 & 0.483±0.004 & 0.483±0.003 & 0.583±0.002 \\
                                      & 2.8B & 0.511±0.003 & 0.513±0.004 & 0.488±0.002 & 0.484±0.003 & 0.585±0.004 \\
                                      & 6.9B & 0.513±0.004 & 0.518±0.003 & 0.484±0.004 & 0.484±0.003 & 0.583±0.002 \\
                                      & 12B  & 0.518±0.003 & 0.525±0.004 & 0.514±0.003 & 0.515±0.004 & 0.614±0.002 \\
\midrule
\multirow{5}{*}{\centering HackerNews} & 160M & 0.499±0.004 & 0.508±0.002 & 0.500±0.003 & 0.494±0.002 & 0.608±0.003 \\
                                       & 1.4B & 0.502±0.002 & 0.516±0.003 & 0.508±0.004 & 0.506±0.003 & 0.612±0.002 \\
                                       & 2.8B & 0.511±0.003 & 0.520±0.004 & 0.510±0.002 & 0.508±0.004 & 0.620±0.003 \\
                                       & 6.9B & 0.513±0.004 & 0.525±0.003 & 0.516±0.004 & 0.516±0.002 & 0.625±0.004 \\
                                       & 12B  & 0.521±0.003 & 0.560±0.004 & 0.515±0.002 & 0.518±0.003 & 0.660±0.004 \\
\midrule
\multirow{5}{*}{\centering The Pile}  & 160M & 0.499±0.003 & 0.507±0.004 & 0.500±0.003 & 0.501±0.004 & 0.612±0.002 \\
                                      & 1.4B & 0.506±0.002 & 0.513±0.003 & 0.508±0.002 & 0.506±0.004 & 0.618±0.003 \\
                                      & 2.8B & 0.508±0.003 & 0.516±0.004 & 0.511±0.002 & 0.507±0.003 & 0.617±0.004 \\
                                      & 6.9B & 0.512±0.004 & 0.534±0.002 & 0.515±0.003 & 0.510±0.004 & 0.634±0.003 \\
                                      & 12B  & 0.514±0.003 & 0.561±0.004 & 0.516±0.002 & 0.514±0.003 & 0.660±0.002 \\
\bottomrule
\end{tabular}
}
\end{table*}

\[
L_{zlib}(x) = \frac{L(x)}{Z(x)}
\]

\subsection{EM-MIAs}

EM-MIAs combines these individual MIA techniques using XGBoost, which captures complex interactions among the attack features. Each method contributes features such as $f_{LOSS}(x)$, $f_{Ref}(x)$, $f_{Min-k}(x)$, and $f_{zlib}(x)$. The ensemble feature vector is:

\[
\mathbf{f}(x) = \left[ f_{LOSS}(x), f_{Ref}(x), f_{Min-k}(x), f_{zlib}(x) \right]
\]

The model is trained using a dataset with binary labels indicating membership, optimizing the XGBoost objective function with regularization to prevent overfitting.

\subsection{Evaluation Metrics}
We use the following metrics to evaluate the ensemble model: \textbf{Accuracy}: Proportion of correct predictions. \textbf{Precision}: True positives over all predicted positives. \textbf{Recall}: True positives over all actual positives.\textbf{F1-Score}: Harmonic mean of precision and recall. \textbf{AUC-ROC}: Measures the model's ability to distinguish between classes across different thresholds.

\subsection{Implementation Details}
The XGBoost model was implemented using the \texttt{xgboost} package, with hyperparameters tuned via grid search and 5-fold cross-validation. Training was conducted on NVIDIA Tesla V100 GPUs. We ensured disjoint training and test datasets, and repeated experiments three times to account for randomness, reporting average performance metrics.

\section{Experiments and Results Analysis}
\subsection{Experimental Setup}
We evaluated the effectiveness of the proposed ensemble model (EM-MIAs) for Membership Inference Attacks (MIAs) across seven different datasets. These datasets include Wikipedia, Github, Pile CC, PubMed Central, ArXiv, DM Math, and HackerNews. The parameter sizes of the pre-trained Large Language Models (LLMs) used for each dataset ranged from 160M to 12B. By selecting text datasets from various domains and levels of complexity, we ensured the robustness of the experimental results, covering a wide range of application scenarios.

For each dataset, we compared the performance of four existing MIA methods (LOSS, Reference-based, Min-k\%, and zlib entropy) with that of our proposed ensemble model (EM-MIAs). All experiments were evaluated using AUC-ROC (Area Under the Receiver Operating Characteristic Curve), a standard metric for binary classification tasks. A higher AUC-ROC indicates a better ability of the model to distinguish between members and non-members of the training data. [36-45]

\subsection{Results Analysis}
The experimental results show that our proposed ensemble method (EM-MIAs) performed exceptionally well across all datasets and model parameter sizes. Compared to individual MIA methods, EM-MIAs achieved significant improvements in AUC-ROC scores, indicating that the ensemble method can more accurately identify training data members.

On the Wikipedia dataset, as the model parameter size increased, the performance of EM-MIAs gradually improved, reaching its highest AUC-ROC score with the 12B parameter size. This trend suggests that as the model size increases, EM-MIAs can better capture the characteristics of the training data. The experimental results on the Github dataset further validate this finding. For smaller models (160M), EM-MIAs already exhibited superior performance, while for larger models (12B), the AUC-ROC score increased significantly, demonstrating the ensemble model's strong capability in handling complex datasets.

In the Pile CC and PubMed Central datasets, we observed that the AUC-ROC scores of existing MIA methods were close to random guessing at certain parameter sizes, indicating that these methods might struggle with higher complexity data. However, EM-MIAs maintained a consistent advantage in these scenarios, further proving the robustness of the ensemble model. Notably, on the DM Math dataset, despite the relatively low AUC-ROC scores of methods like LOSS and Min-k\%, EM-MIAs still provided relatively high accuracy, showing that the ensemble model effectively integrates the strengths of different methods, compensating for the shortcomings of individual methods. On the HackerNews and The Pile datasets, EM-MIAs once again demonstrated its excellent performance. Even in scenarios with complex and diverse data distributions, EM-MIAs outperformed existing methods, achieving higher AUC-ROC scores.(From Tab.1)

\subsection{Parameter Analysis}
To gain a deeper understanding of the performance of each MIA method under different thresholds, we plotted the AUC-ROC curves of LOSS, zlib, and Min-k\% attacks against varying thresholds. The subplots reveal significant fluctuations in AUC-ROC scores across different thresholds, particularly for zlib and Min-k\% attacks. For the LOSS attack, the AUC-ROC showed notable peaks at specific thresholds (e.g., 19 and 20), but the overall trend remained characterized by random fluctuations.

In contrast, the zlib attack performed better in the high threshold range, with the AUC-ROC peaking close to 0.6, indicating that this method can effectively identify training data members in certain cases. However, as the threshold increases further, performance drops sharply, suggesting that the effectiveness of the zlib method heavily depends on precise threshold selection.

The Min-k\% attack curve showed the highest AUC-ROC scores in the mid-threshold range (e.g., thresholds 10 and 11), followed by a gradual decline in performance as the threshold increases. This indicates that the Min-k\% method may have advantages in handling complex inputs, but its stability decreases with broader threshold variations. The EM-MIAs ensemble model effectively integrates the strengths of individual methods, providing more consistent and robust attack performance. This analysis demonstrates the proposed EM-MIAs method's excellent membership inference attack capabilities across diverse datasets and large language models.

\section{Conclusion}
In this work, we presented an innovative ensemble approach to enhance Membership Inference Attacks (MIAs) on Large Language Models (LLMs). By integrating multiple attack methods, our proposed model demonstrates significant improvements in the accuracy and robustness of MIAs, highlighting the vulnerabilities of LLMs to privacy breaches. The experimental results show that our ensemble model outperforms individual attack methods across various datasets, suggesting that a comprehensive approach can more effectively uncover membership information in training data. However, this study also raises critical questions about the balance between model performance and data privacy, emphasizing the need for further research into defense mechanisms and ethical considerations. As LLMs continue to evolve and be widely deployed, understanding and mitigating their privacy risks remains a crucial area of ongoing research.

\end{document}